\documentclass[sigplan,screen]{acmart}

\AtBeginDocument{%
  }

\copyrightyear{2023} 
\acmYear{2023} 
\setcopyright{acmlicensed}\acmConference[FDG 2023]{Foundations of Digital Games 2023}{April 12--14, 2023}{Lisbon, Portugal}
\acmBooktitle{Foundations of Digital Games 2023 (FDG 2023), April 12--14, 2023, Lisbon, Portugal}
\acmPrice{15.00}
\acmDOI{10.1145/3582437.3587206}
\acmISBN{978-1-4503-9855-8/23/04}

\usepackage{caption}
\usepackage{subcaption}
\usepackage{hyperref}

\begin{document}

\title{Lode Enhancer: Level Co-creation Through Scaling}


\author{Debosmita Bhaumik}
\affiliation{%
  \institution{Institute of Digital Games}
  \streetaddress{20 Triq L-Esperanto}
  \city{Msida}
  \country{Malta}}
\email{debosmita.bhaumik01@gmail.com}

\author{Julian Togelius}
\affiliation{%
  \institution{Game Innovation Lab, modl.ai}
  \streetaddress{370 Jay Street}
  \country{USA, Denmark}}
\email{julian@togelius.com}

\author{Georgios N. Yannakakis}
\affiliation{%
  \institution{Institute of Digital Games, modl.ai}
  \streetaddress{20 Triq L-Esperanto}
  \country{Malta, Denmark}}
\email{georgios.yannakakis@um.edu.mt}

\author{Ahmed Khalifa}
\affiliation{%
  \institution{Institute of Digital Games}
  \streetaddress{20 Triq L-Esperanto}
  \city{Msida}
  \country{Malta}}
\email{ahmed@akhalifa.com}

\renewcommand{\shortauthors}{Bhaumik et al.}

\begin{abstract}
We explore AI-powered upscaling as a design assistance tool in the context of creating 2D game levels. Deep neural networks are used to upscale artificially downscaled patches of levels from the puzzle platformer game \emph{Lode Runner}. The trained networks are incorporated into a web-based editor, where the user can create and edit levels at three different levels of resolution: 4x4, 8x8, and 16x16. An edit at any resolution instantly transfers to the other resolutions. As upscaling requires inventing features that might not be present at lower resolutions, we train neural networks to reproduce these features. We introduce a neural network architecture that is capable of not only learning upscaling but also giving higher priority to less frequent tiles. To investigate the potential of this tool and guide further development, we conduct a qualitative study with 3 designers to understand how they use it. Designers enjoyed co-designing with the tool, liked its underlying concept, and provided feedback for further improvement. 
\end{abstract}


\begin{teaserfigure}
  \centering
  \includegraphics[width=0.75\textwidth]{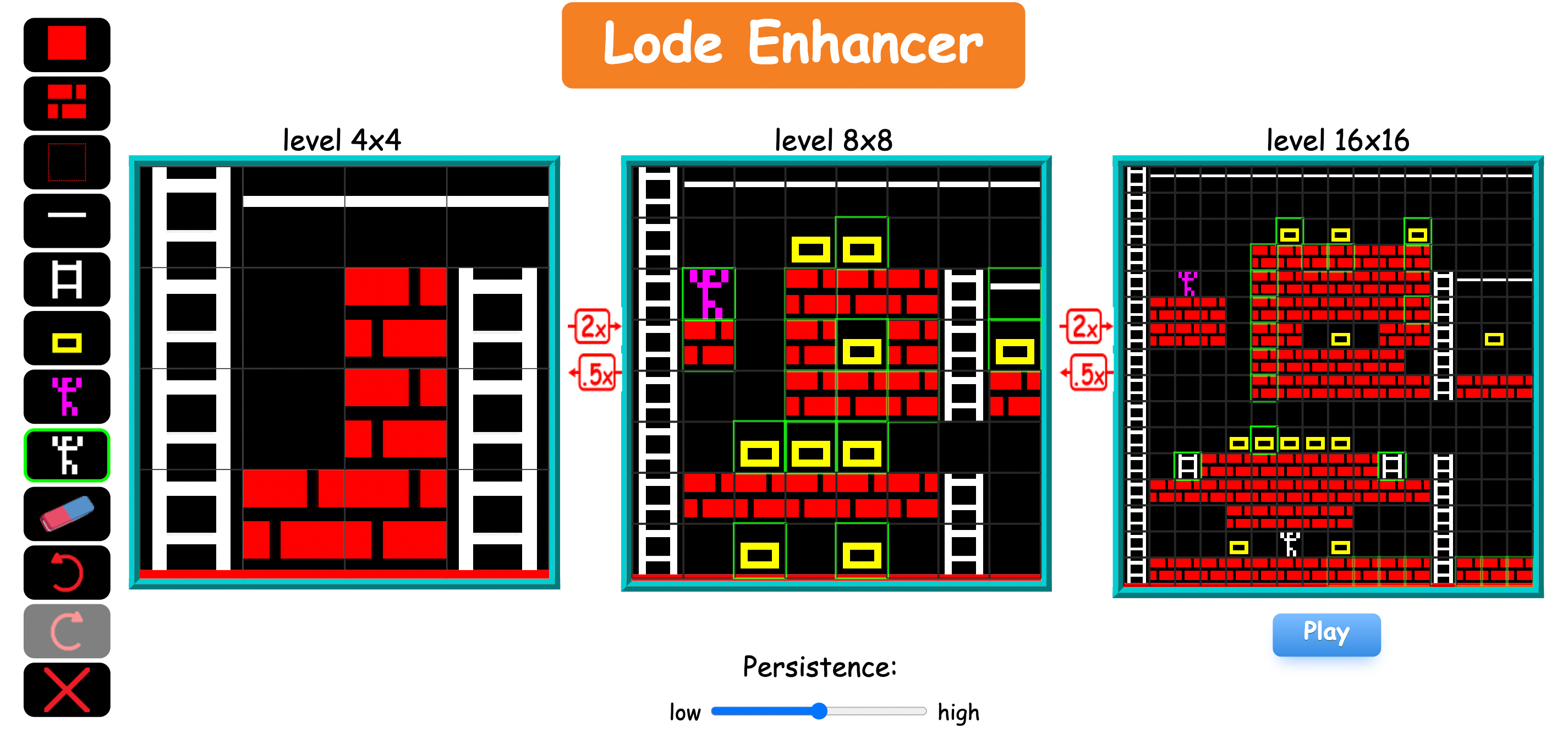}
  \caption{Lode Enhancer user interface with an example of an up-scaled level. The system is always running whenever a user makes a change in any canvas the other two get updated. 
  The persistence bar helps to inform the system on which tiles can be replaced during scaling.
  }
  \label{fig:system_ui}
\end{teaserfigure}


\maketitle

\section{Introduction}

Artificial Intelligence (AI)-powered design assistance can take many forms, thereby, experimenting with the relation between human user(s) and AI systems is crucial for establishing human-AI co-creativity. A common trope is that of the human designing something relatively small or lacking detail (i.e. \emph{downscaled} design), and the AI system producing a larger and more detailed (i.e. \emph{upscaled}) version of the human design~\cite{tropes2007enhance}. This entails that the AI system has some way of making up for the missing information. 

In this paper, we explore the idea of using AI-powered upscaling (and, to a lesser extent, downscaling) for design assistance. The core idea is that a designer can design at any level or resolution (i.e. scale) and the AI system will intelligently up- and downscale it as necessary. In particular, a designer can draw a game-level sketch at a small scale and have it upscaled to a complete level, and then edit the upscaled level which gets automatically scaled down so it reflects the complete level. 

The prototype system we describe in this paper, \emph{Lode Enhancer}, uses a neural architecture to scale up level sketches from 4x4 to 8x8 tiles, and again from 8x8 to 16x16 tiles (figure~\ref{fig:system_ui}). The neural networks are trained to upscale downscaled versions of existing Lode Runner level segments, meaning they have learned common Lode Runner level design patterns. The complete system, including trained networks, runs in the browser and produces near-instant results on modern computers. To understand the potential of AI-powered upscaling and downscaling, we did a quantitative analysis on the generated levels then followed by an informal study where we let three game designers use \emph{Lode Enhancer} and interviewed them about their impressions. 


\section{Lode Runner}


Lode Runner (Broderbund, 1983)
is a classic puzzle platformer game where  
the goal of the game is to collect all gold pieces without being caught by the enemies. The player can only move left and right and dig through the floor. To go to higher platforms, the player needs to climb ladders. The digging mechanic allows the player to travel downward 
and it can be used to trap enemies. 

Lode Runner comes with 150 levels which makes it a good candidate for procedural level generation using machine learning (PCGML)~\cite{summerville2018procedural}. Also, the spatial dependencies between different tiles make this game a good test bed for our experiment. Snodgrass and Ontan{\'o}n~\cite{snodgrass2016learning} used multidimensional markov chain to generate levels for different games including Lode Runner. Thakkar et al.~\cite{thakkar2019autoencoder} used a trained vanilla and variational autoencoder to generate Lode Runner levels and used evolution strategies to search for playable levels. 
Steckel and Schrum~\cite{steckel2021illuminating} trained a GAN on Lode Runner levels and used the MAP-Elites algorithms to search the space for diverse playable levels. 

\section{Mixed-Initiative PCG}


There exists various literature on procedural content generation,
in particular on the autonomous generation of levels where the system produces levels with minimal human input~\cite{shaker2016procedural}. In many settings, it is more useful to have a system that can interact with humans, so that a human user and an AI system can design together~\cite{liapis2016mixed,yannakakis2014mixed}. Many of these systems are based on making suggestions to the user, evaluating the user's edits, and/or enforcing constraints of various kinds. For example, Tanagra~\cite{smith2011tanagra} uses constraint solving to guarantee playability in user designs, and Sentient Sketchbook evaluates strategy maps for various kinds of balance and uses evolutionary algorithms to suggest changes to the maps~\cite{liapis2013sentient}. 

While most early mixed-initiative game design assistants relied on search, optimization, and/or constraint solving, a new generation of these tools build on machine learning. For example, RL brush~\cite{delarosa2021rlbrush} gives level editing suggestions to users generated by reinforcement learning agents, and Lode Encoder~\cite{bhaumik2021lode} allows users to mix and match between suggested levels using variational autoencoders and incentivizes users to create new levels. 
In Morai Maker~\cite{guzdial2019morai}, the human user and the AI agent collaborate in a turn-based process to generate Super Mario Bros (Nintendo, 1985) levels. 


\begin{figure*}[!tb]
    \centering
    \includegraphics[width=0.9\textwidth]{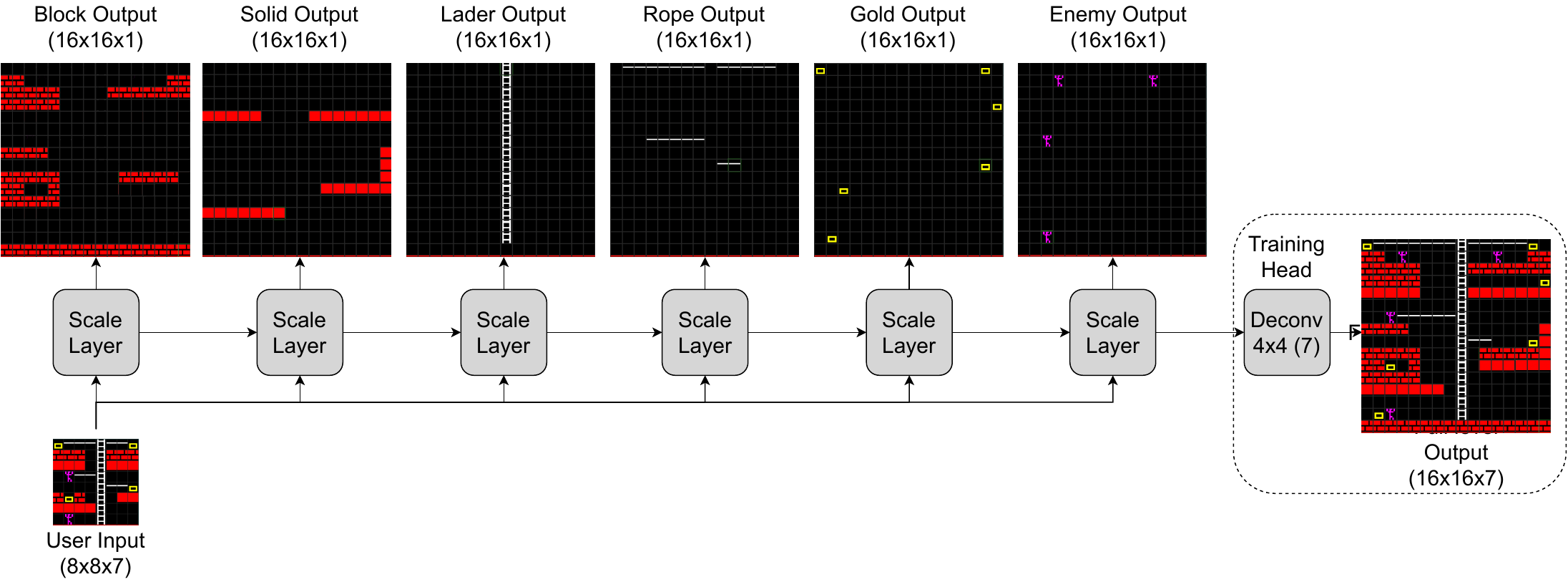}
    \caption{The layer scaling network architecture. The network consists of 6 scale layers (one for each tile type) followed by a temporary Deconv layer to help train the network. The input for the first scale layer is ignored and only the user input is used.}
    \label{fig:network_arch}  
\end{figure*}

\section{Multilevel Scale Machine Learning}

A staple of sci-fi movies and crime shows, image upscaling~\cite{tropes2007enhance}, is now a commonplace technique that is used both on its own~\cite{dong2014learning} and as part of image generation workflows~\cite{rombach2022high}. Neural networks of various types can be trained to upscale images simply by using datasets where an artificially downscaled image is the input and the original image is the target. As the network learns to upscale, it learns to reproduce the various aspects of a high-resolution image that is not part of its low-resolution counterpart.

Upscaling is commonly used in image generation pipelines. This includes the StyleGAN~\cite{karras2020analyzing} family of networks, where original images are generated at a low resolution and then upscaled by successive networks. 
Similarly, Stable Diffusion~\cite{rombach2022high} uses trained upscaling diffusion models to generate high-quality images. Upscaling has not been applied much to mixed-initiative game-level generation yet. The most closely related work is Sentient World~\cite{liapis2013sentientworld}, where the user designs a low-resolution map, then the system suggests an upscaled version using neuroevolution through novelty search.  
In Sentient World, the generated maps are a continuous value for the height map which makes the problem similar to images compared to generating levels for tile-based games where there is no correlation between the tile values.


Game level design poses a significantly different problem than image generation, because of the functionality criteria; an image does not need to ``work'', but a level needs to be completable. 
For levels based on fixed-size assets (such as tiles/voxels), there is also the phenomenon that this functionality constraint might only exist at the full-resolution version of the level, as the compressed levels are almost not completable. 
Another common challenge is the discrete aspect of the domain. When you scale up a level it does not mean that the surrounding tiles/voxels suppose to be similar. For example, having one enemy in a low-resolution level does not mean that we need 10 enemies at this position if the level got enlarged 10 times.

\section{Lode Enhancer}

Lode Enhancer\footnote{\url{http://www.akhalifa.com/lodeenhancer/}} is an AI-powered level design tool that helps game designers to create ``Lode Runner'' levels through upscaling. Using the system, the user can draw levels on a small canvas then the system upscales it in an intelligent way and produces a larger and more detailed level. Figure~\ref{fig:system_ui} shows the full system UI where the user is provided with three different canvases of different scales (4x4, 8x8, and 16x16). The user has a toolbar on the left that can be used to pick a certain tile to draw. Whenever any change happens to any canvas, the canvas is sent to the scaler module which reflects these changes to the other two canvases. Finally, the scaler module uses the persistence module (shown as a slider at the bottom of the UI) to know which tiles can be replaced and which ones should be kept the same. 

\subsection{Scaler Module}

The scaler module is responsible for reflecting any changes that the user makes in any canvas to the other canvases. This is done using an up-scaler system and a down-scaler system. The up-scaler system is responsible for doubling the level size while the down-scaler system is responsible for halving the level size. For example, if the user changes the 8x8 canvas, the system will use the down-scaler to reflect these changes in the 4x4 canvas and the up-scaler to reflect the changes in the 16x16 canvas.

The up-scaler system is modeled using a neural network called layer scaling network (explained in section~\ref{sec:scale_network}), while the down-scaler uses a traditional nearest neighbor filter to shrink the level. Originally, both systems were modeled using neural networks but in early experiments, we discovered that the down-scaler network learns to almost replicate the nearest neighbor filter. 

\subsection{Persistence Module}

In our early experiments, we found that the user's edits in the 16x16 canvas got overwritten by the scaler module if the user went back and modified the 8x8 or 4x4 canvas. This problem prevented a lot of early testers from going back and modifying the 8x8 or 4x4 canvas after making edits to the 16x16 canvas. To solve this, we introduced the persistence slider that can be seen at the bottom of figure~\ref{fig:system_ui}. The persistence slider tells the AI how much it should respect the user's edits. For high persistence, the user's edits get the highest priority and shouldn't be replaced. While in low persistence, the AI can overwrite easily any tile. 

The module calculates a confidence value (between 0.5 and 1.0) for each user-drawn tile. To replace a tile, the scaler module compares the probability of the tile from the neural network with its confidence value. The tile gets replaced if the network's probability is higher than the confidence value of the same tile. To calculate this confidence value, we take into account the age of the tile (i.e. how long ago the tile was drawn) as we want the older tiles to be easier to replace than newly drawn tiles. Equation~\ref{eq:persistence} shows how the confidence value ($C$) is calculated given the age of the tile ($a$) where $C_{max}$ is the maximum confidence value, $a_{min}$ is the age at which the tile confidence start decreasing linearly, and $a_{max}$ is the age at which the tile can be replaced.
\begin{equation}\label{eq:persistence}
    C(a) = \begin{cases}
        C_{max} & a \leq a_{min}\\
        \frac{C_{max} - 0.5}{a_{max} - a_{min}} \cdot (a_{max} - a) + 0.5 & a_{min} > a \leq a_{max}\\
        0.5 & a > a_{max}
    \end{cases}
\end{equation}
Moving the persistence bar changes the $a_{min}$, the $a_{max}$, and the $C_{max}$ values. For the lowest persistence $a_{min}$ is $0$, $a_{max}$ is $1$, and $C_{max}$ is $0.5$, while for the highest persistence $a_{min}$ is $20$, $a_{max}$ is $100$, and $C_{max}$ is $1$. The slider just linearly interpolates these values for the in-between ticks.

\section{Layer Scaling Network}\label{sec:scale_network}

Layer scaling network is a new network architecture (shown in figure~\ref{fig:network_arch}) designed to help with discrete domains with uneven distribution between the different possible values/tiles. This new architecture gives different priorities and computation capacities for the different tiles. 
For example, in Lode Runner, we want the gold tiles and enemy tiles to have higher priority than normal brick tiles. 

To achieve this, our network consists of $n-1$ scale layers (explained in section~\ref{sec:scale_layer}) where $n$ is the number of possible tiles. Each layer takes the main user input and outputs a probability of a specific tile. The order of the output is picked based on the rarity of each tile. For example, in Lode Runner, the enemy tile (rarest) becomes the last layer while the brick tile (most common) is the first layer. For Lode Runner (see figure~\ref{fig:network_arch}), the network consists of six scale layers: one layer for each tile type except for empty and player tiles. We removed the player tile as we wanted the user to have control over the starting location, while the empty tile is the default tile if nothing overwrites the location.

\begin{figure}[!tb]
    \centering
    \includegraphics[width=0.9\linewidth]{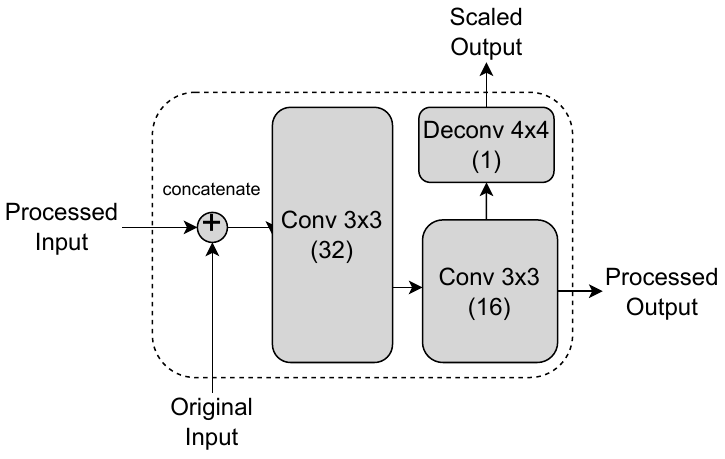}
    \caption{The scale layer used in the new network architecture. The layer takes two inputs (the previous layer output and the user input) and outputs the scaled output 
    and 16-channel output for network connectivity. 
    }
    \label{fig:scale_layer}
\end{figure}

\subsection{Scale Layer}\label{sec:scale_layer}

The scale layer (as shown in figure~\ref{fig:scale_layer}) consists of 2 convolution layers and a deconvolution layer. 
The layer takes two inputs: the original input and the processed input and produces two outputs: the scaled output and the processed output. The original input is the small-scale user-drawn level, this input is always concatenated to the processed input so the layer is conditioned by the user's choices. This is similar to the skip connections in the ResNet~\cite{he2016deep}. 
Both inputs are of the same size but have a different number of channels. The scaled output, instead, is twice the size of the input with one channel where this output reflects the probability of a certain tile type. The processed output is the normal output that gets pushed toward the next layer as its processed input. 

For Lode Runner, we used two convolutional layers with ReLu activation function followed by a batch normalization layer for the processed output. The scaled output uses a deconvolution layer to scale the input dimension to a bigger dimension. We did not add more layers as the dataset size is not big enough and we did not want to give the network more power to easily overfit.

\subsection{Network Feedforward Operation}

The user-drawn level (small-scale discrete image with 7 channels) is fed to the network and we collect the different scaled outputs from the scale layers (6 different outputs). To interpret the output, we first create a level of twice the size and set it to empty tiles. We go over the scaled output from the beginning (brick layer) to the end (enemy layer). For each scaled output, if the value is greater than $0.5$, we change the tile to the current layer tile, otherwise, we leave it as it is. For example, if we are checking the ladder scaled output and the value of the tile at (x=0, y=0) is more than $0.5$, we change the tile in the final output at (x=0, y=0) to the ladder tile. If later at the gold scaled output that same tile location has a value of more than $0.5$, the tile is changed to gold instead. 

\subsection{Network Training}

To train this architecture, we apply a two-step training method: base training and greedy layer training. The base training adds a deconvolution layer (called training head in figure~\ref{fig:network_arch}) after the last scale layer. The deconvolution layer produces a scaled output with 7 channels using a softmax activation function (representing all the different tile values). We use Adam optimizer 
to train the network using categorical cross-entropy loss function for $3,000$ epochs. 

After the base training mentioned above, we run greedy layer training to fine-tune each scale layer and train their deconvolution layer. We start by removing the training head and go in order of layers from the first scale layer to the last. We get the corresponding channel from the training data and fine-tune each layer. We use Adam optimizer to fine-tune the layer using binary cross-entropy loss function for $1,000$ epochs with early stopping. After a layer is fine-tuned, we freeze its weights and move to the next layer until all layers are fine-tuned. For Lode Enhancer, we trained two different models: scale from 4x4 to 8x8 and scale from 8x8 to 16x16. 

\begin{figure}[!tb]
    \centering
    \begin{subfigure}[t]{0.49\linewidth}
        \centering
        \includegraphics[width=0.9\textwidth]{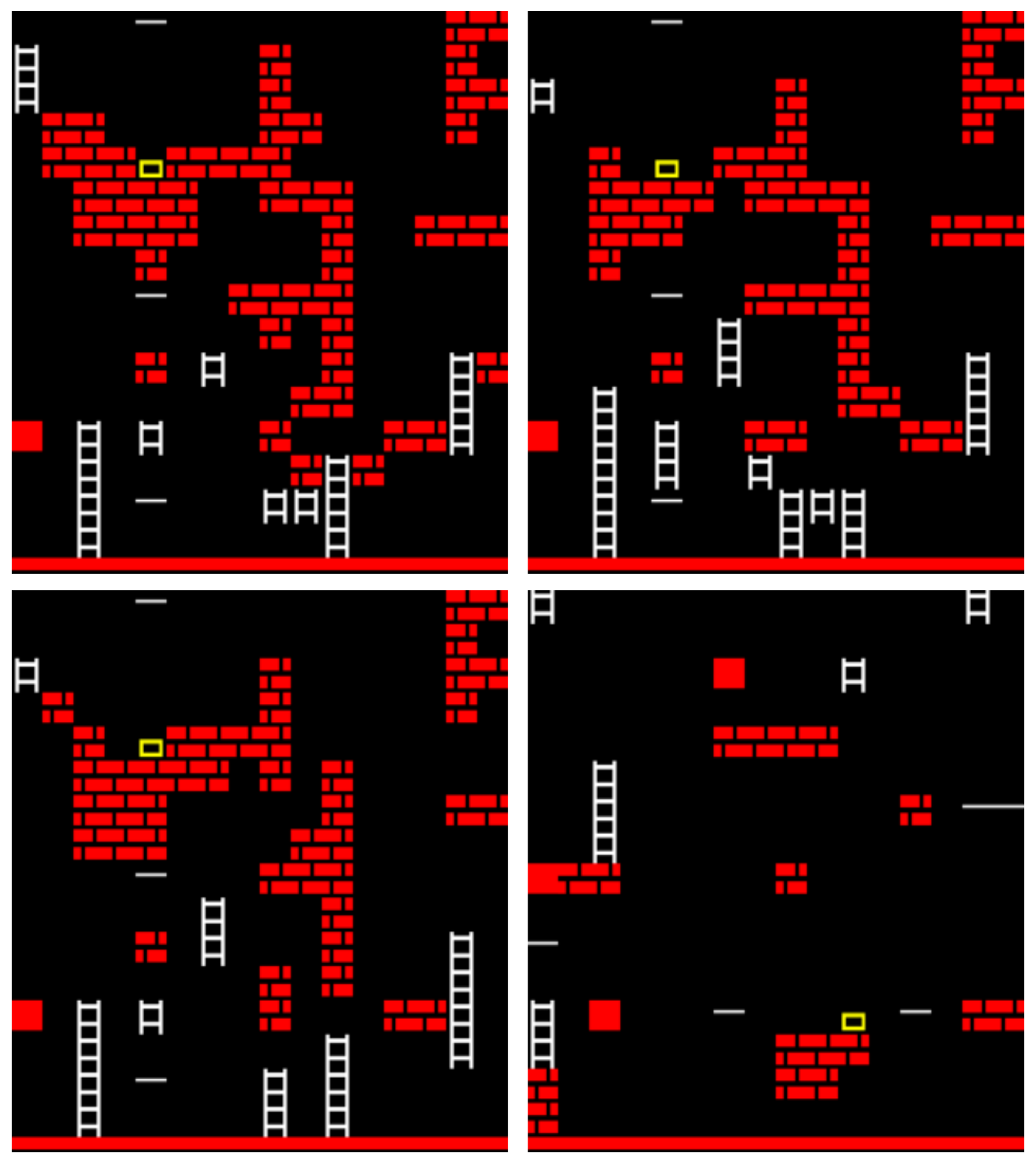}
        \caption{8x8 starting noise}
        \label{fig:pipe8_lvls}
    \end{subfigure}
    \begin{subfigure}[t]{0.49\linewidth}
        \centering
        \includegraphics[width=0.9\textwidth]{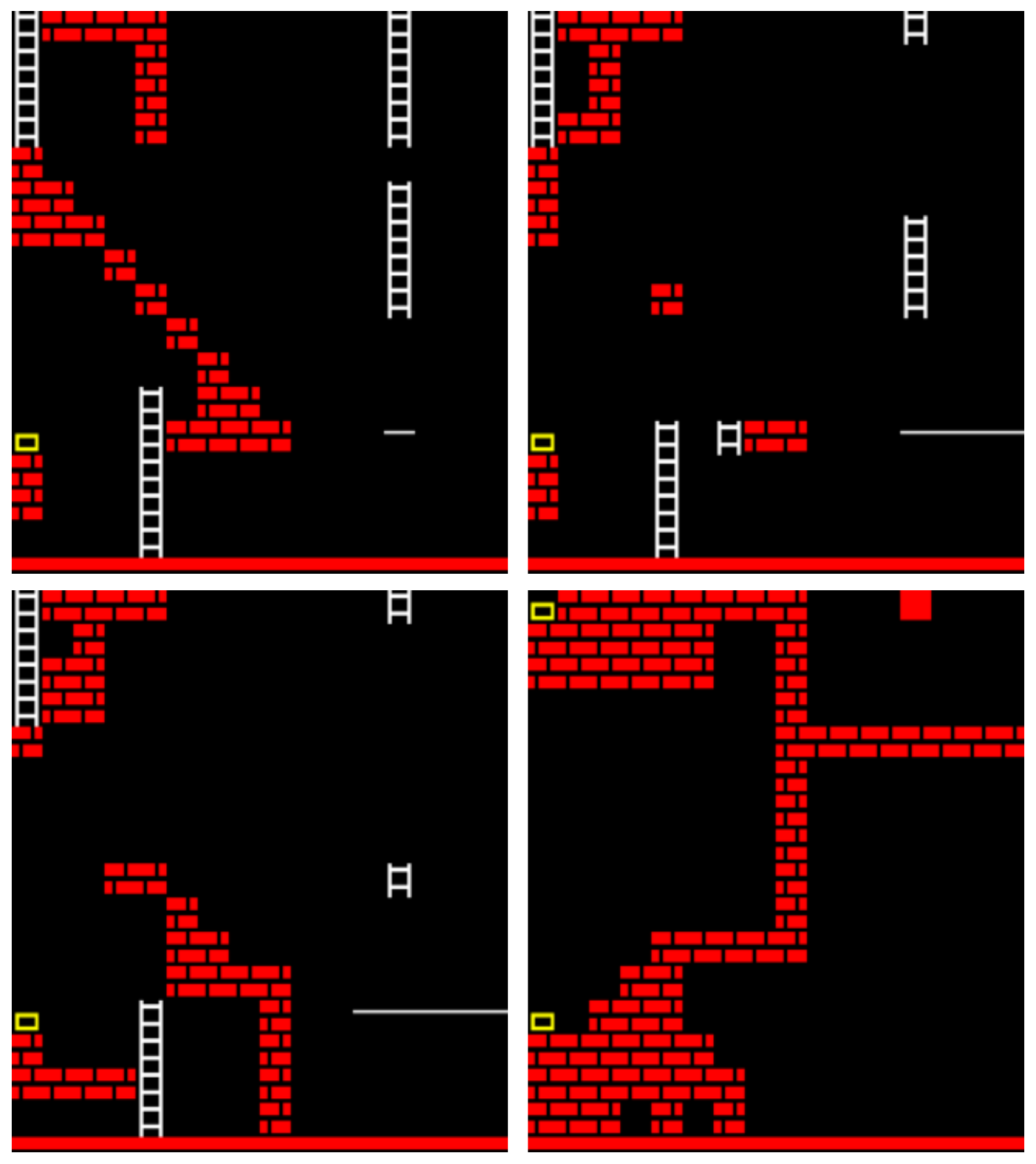}
        \caption{4x4 starting noise}
        \label{fig:pipe16_lvls}
    \end{subfigure}
    \caption{Examples of scaled levels from different noise sizes.}
    \label{fig:scale_levels}
\end{figure}

\subsection{Dataset}\label{sec:dataset}

For our experiment we used 150 Lode Runner levels made of 32x22 tiles from the Video Games Level Corpus (VGLC)~\cite{summerville2016vglc}. To increase the dataset size, we used smaller segments of original levels. We settled down for 4x4, 8x8, and 16x16. 
For our first network that scales 8x8 levels to 16x16, we applied a 16x16 sliding window method with a stride of 1 which produced $17,850$ level segments in total. Since Lode Runner levels can be reflected across the x-axis, we used that method to augment our dataset to $35,700$ level segments. We then scaled down these segments to half of their size (8x8) using the nearest neighbor filter and paired them together to create the dataset. We repeated the same process using a sliding window of 8x8 to create a dataset for our second network that scales up from 4x4 to 8x8. The dataset for the second network consists of $112,500$ level segments.

\section{Layer Scaling Network Analysis}

In this section, we analyze quantitatively our layer scaling network. We focus on two fronts: a) investigating the effect of network scaling on the input level and b) investigating the advantages of using the proposed architecture.


\subsection{Scaling Effect}

In this subsection, we explore the effect of the network's scaling.
We started by generating $1,000$ 4x4 or 8x8 noise inputs that follow the same distribution of the training data. We passed the 8x8 inputs to 3 trained networks to scale the 8x8 to 16x16, giving us $3,000$ 16x16 levels. We passed the $1,000$ 4x4 inputs through 3 trained networks to scale the 4x4 to 8x8 and then passed the outputs through the other network to scale it to 16x16, giving us $3,000$ 16x16 levels. 

\begin{figure}[!tb]
    \centering
    \includegraphics[width=0.7\linewidth]{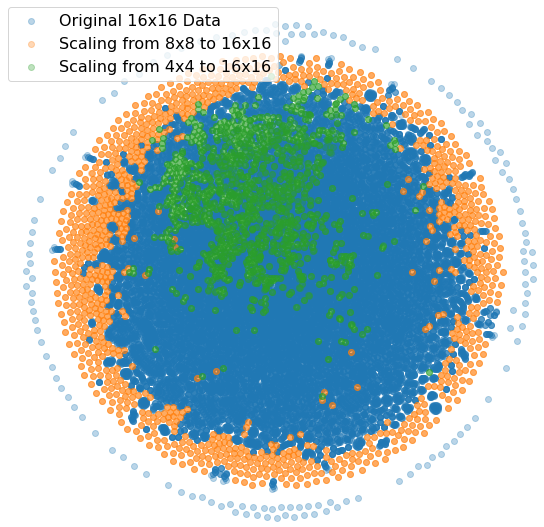}
    \caption{Rendering the training levels with different generated levels from the layer scaling network using t-SNE.}
    \label{fig:tsne}
\end{figure}

Figure~\ref{fig:scale_levels} shows an example of the final scaled levels. It is clear that scaling twice (from 4x4 to 8x8 and then to 16x16) yields better-looking levels compared to scaling once (from 8x8 to 16x16). We believe that having the level iterate more than once in the network helps to smooth the input noise and make the levels fall in a similar distribution to the training data. This can be shown in the t-SNE visualization of all the 16x16 levels (figure~\ref{fig:tsne}).

\subsection{Network Architecture}

\begin{figure*}[!tb]
    \centering
    \begin{subfigure}[t]{0.21\linewidth}
        \centering
        \includegraphics[width=0.9\textwidth]{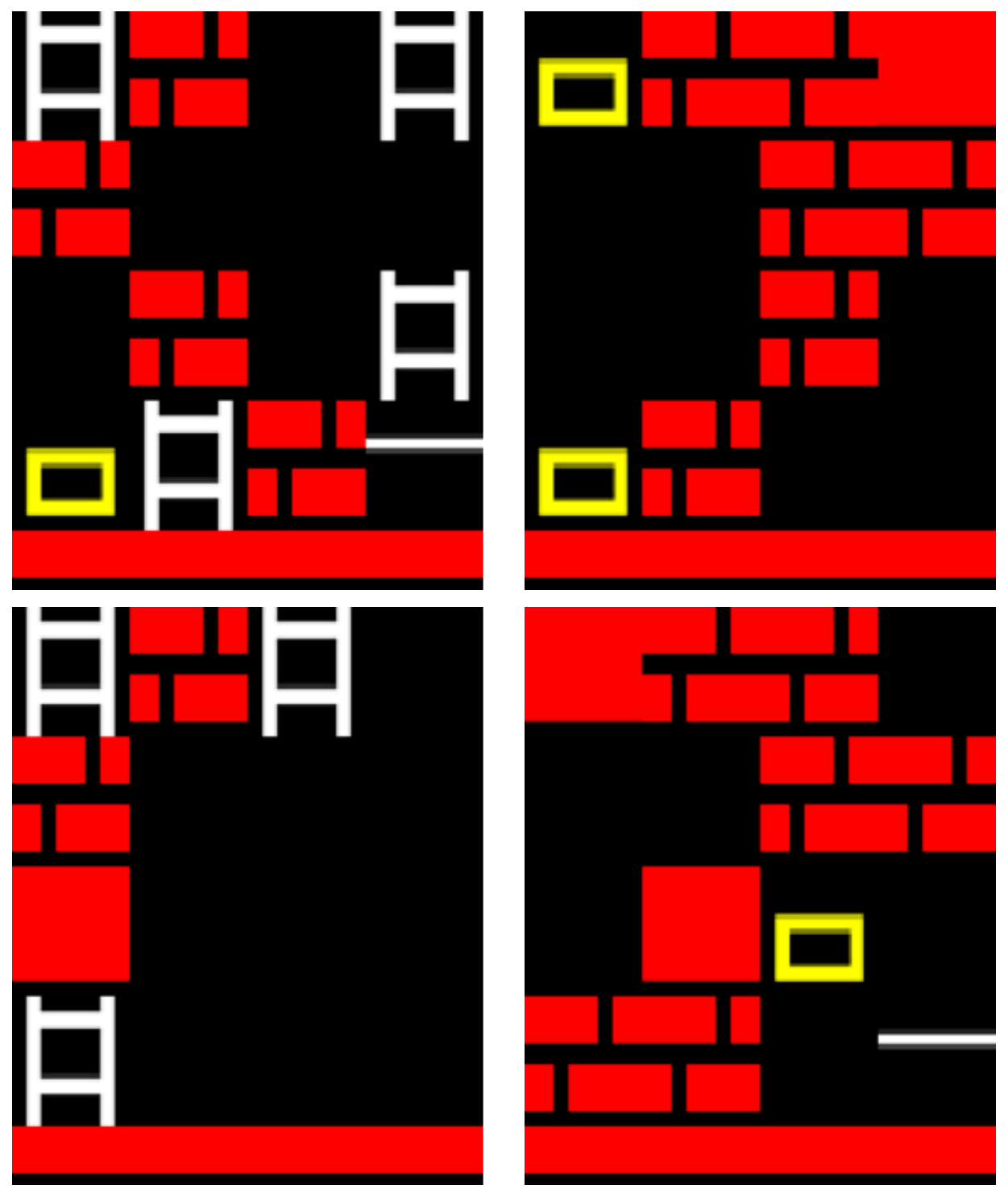}
        \caption{Input Levels}
        \label{fig:input_lvls}
    \end{subfigure}
    \begin{subfigure}[t]{0.21\linewidth}
        \centering
        \includegraphics[width=0.95\textwidth]{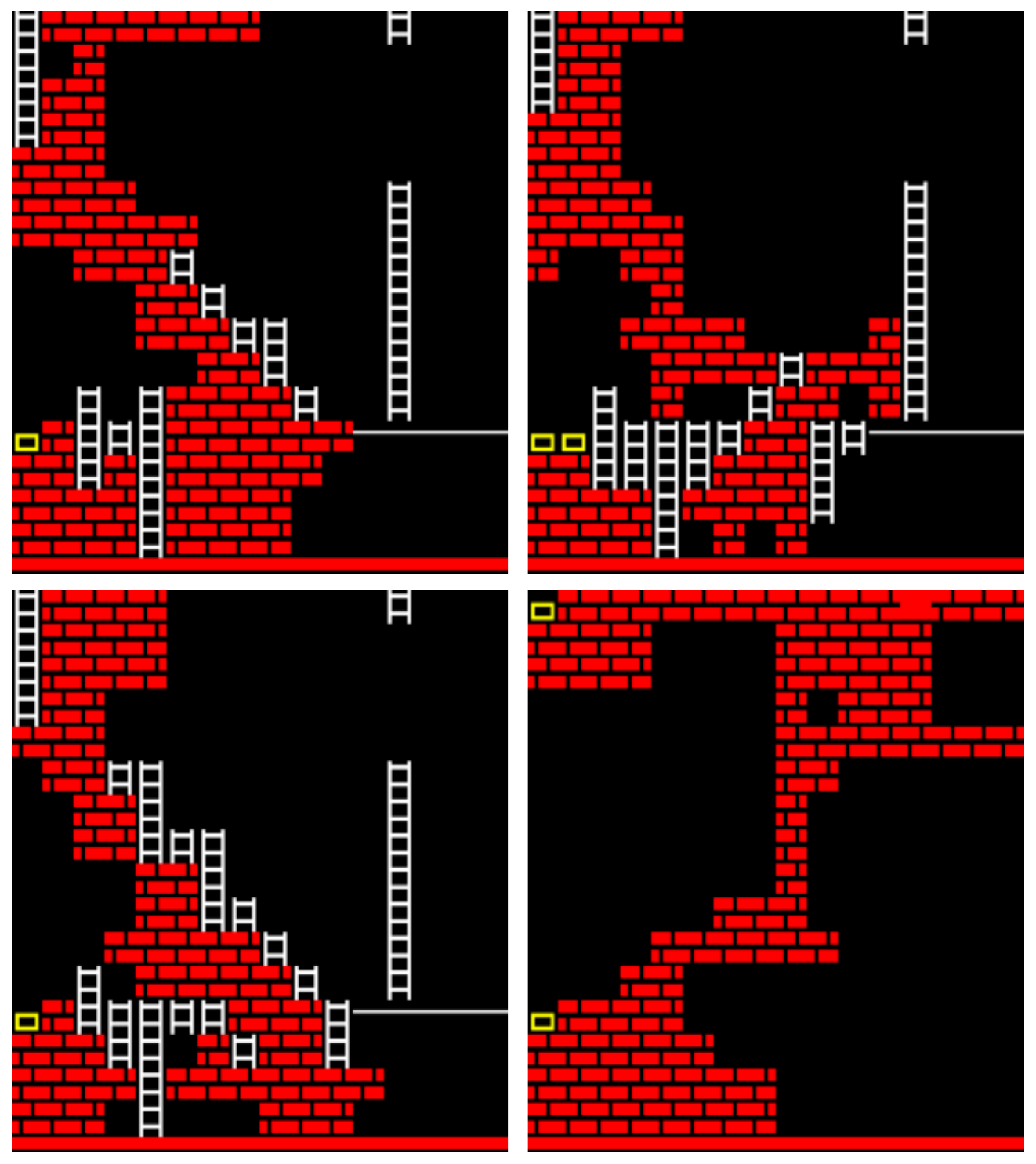}
        \caption{Convolutional Network}
        \label{fig:conv_lvls}
    \end{subfigure}
    \begin{subfigure}[t]{0.21\linewidth}
        \centering
        \includegraphics[width=0.95\textwidth]{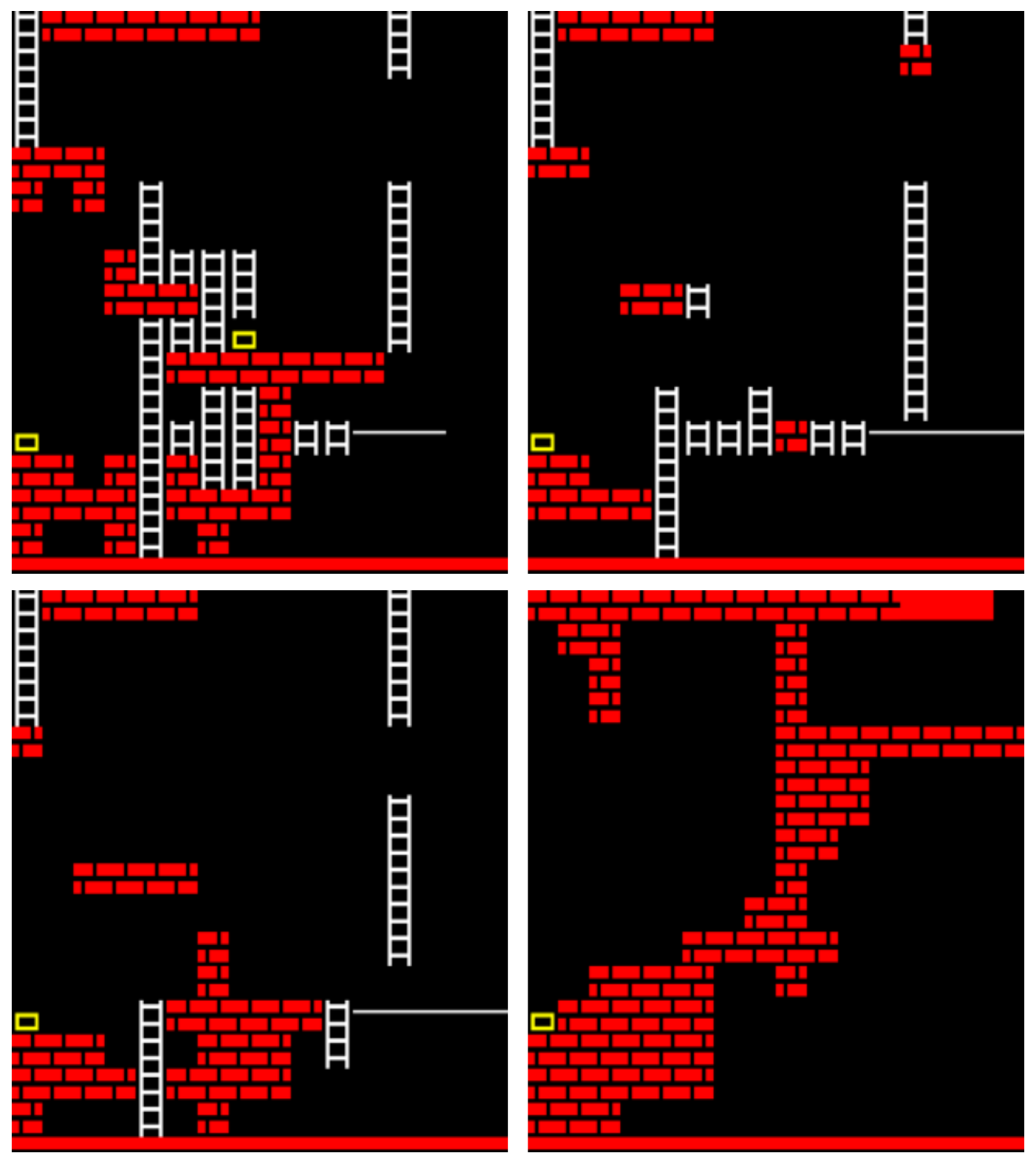}
        \caption{Scaling Network}
        \label{fig:greedy_lvls}
    \end{subfigure}
    \begin{subfigure}[t]{0.21\linewidth}
        \centering
        \includegraphics[width=0.95\textwidth]{images/pipe_levels.pdf}
        \caption{Layer Scaling Network}
        \label{fig:pipe_lvls}
    \end{subfigure}
    \caption{Example of generated 16x16 levels from scaling twice the same input noise using different network architectures.}
    \label{fig:example_levels}
\end{figure*}

\begin{figure*}[!tb]
    \centering
    \begin{subfigure}[t]{0.28\linewidth}
        \centering
        \includegraphics[width=0.95\textwidth]{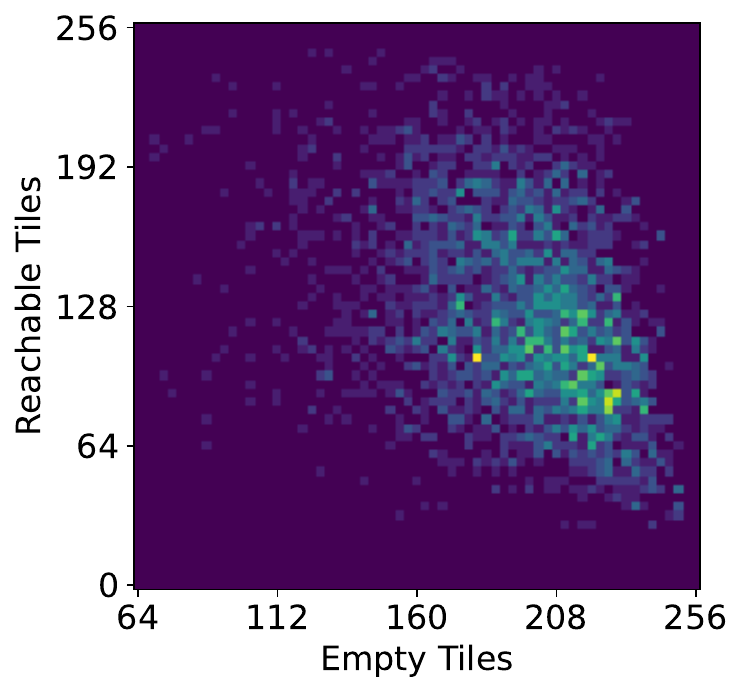}
        \caption{Conv Network}
        \label{fig:conv_range}
    \end{subfigure}
    \begin{subfigure}[t]{0.28\linewidth}
        \centering
        \includegraphics[width=0.95\textwidth]{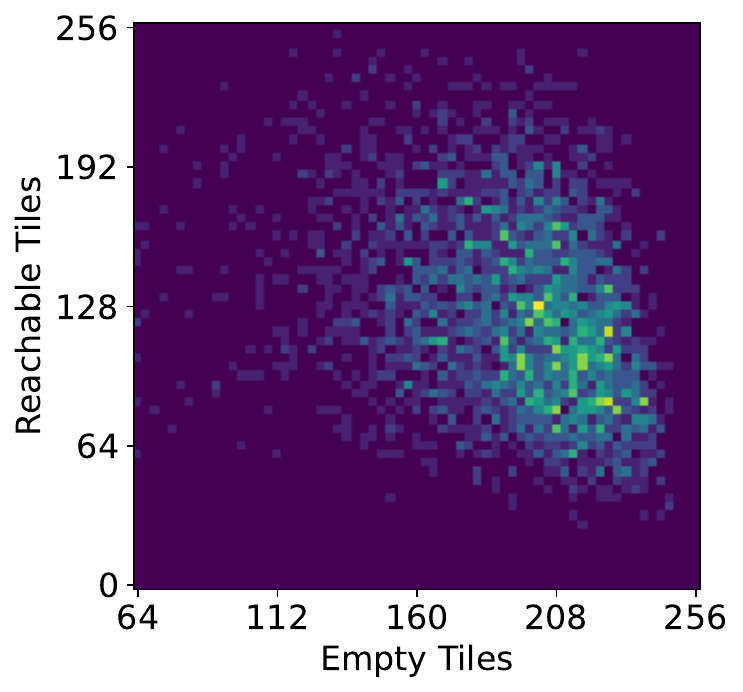}
        \caption{Scaling Network}
        \label{fig:greedy_range}
    \end{subfigure}
    \begin{subfigure}[t]{0.28\linewidth}
        \centering
        \includegraphics[width=0.95\textwidth]{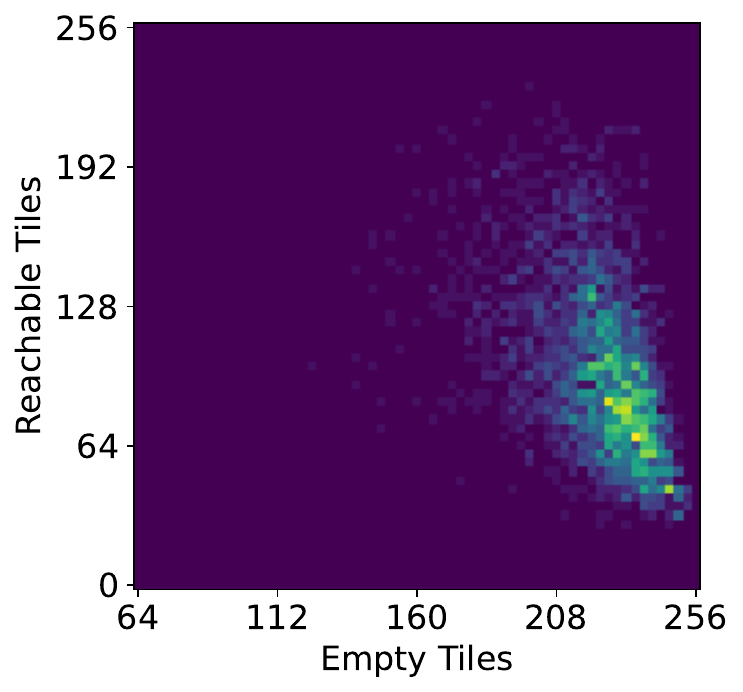}
        \caption{Layer-Scaling Net}
        \label{fig:pipe_range}
    \end{subfigure}
    \caption{Expressive range analysis from scaling twice the same input noise using different network architectures.}
    \label{fig:example_range}
\end{figure*}

In this subsection, we investigate the effect of the proposed network architecture and compare it against a convolutional neural network (i.e. 2 convolutional layers followed by a deconvolutional layer). We also compare the efficiency of the greedy layer training (layer-scaling network) with respect to the training head (scaling network).
For this experiment, we generated $1,000$ levels by scaling up 4x4 noise twice to reach the size of 16x16. This experiment is repeated 3 times using 6 different trained networks (i.e. 3 networks that scale from 4x4 to 8x8 and 3 networks that scale from 8x8 to 16x16) resulting to $3,000$ 16x16 generated levels per experiment. 


We compare the generated levels (figure~\ref{fig:example_levels}) with the original 16x16 training levels and calculate the minimum tile-pattern KL-divergence (TPKLDiv) score~\cite{lucas2019tile}. To calculate the minimum score, we pair every generated level with the closest level from the training levels (i.e. the one with the minimum TPKLDiv score), and we compute the average score and the confidence interval for all the different pairs. We find that both the convolutional network (\textit{3.485 ± 0.084}) and the scaling network (\textit{3.633 ± 0.041}) yield higher scores compared to the layer scaling network (\textit{2.441 ± 0.031}). Having a lower TPKLDiv score is desired as it means that the local patterns in the generated levels follow a similar distribution to the ones from the training levels. 


Figure~\ref{fig:example_range} shows the expressive range analysis for the different networks. We used two main metrics: the number of reachable tiles and the number of empty tiles. The number of reachable tiles is the maximum number of reached tiles by a breadth-first search playing agent that is tested from every single location in the level. Looking at the expressive range, we notice that both convolutional and scaling networks have similar ranges. On the other hand, the layer scaling network yields a more consistent map with a lot of empty tiles and average reachable tiles. This result is unsurprising as having a higher amount of empty tiles reduces the number of reachable tiles since the main character cannot jump. This finding also suggests that many of the generated levels from the layer scaling network are similar to each other, which can be seen in figure~\ref{fig:pipe_lvls}. We believe that this is an advantage to our network as it keeps the relationship between the input (figure~\ref{fig:input_lvls}) and the output (figure~\ref{fig:pipe_lvls}) as close as possible. In turn, this means that a small change in the input level will not cause drastic changes in the output level. This characteristic is particularly important for mixed-initiative tools of higher trustworthiness value, as the user can better understand the impact of their actions on the generated level.

\section{Qualitative evaluation}

To evaluate Lode Enhancer and the ideas behind it, and to provide directions for future development, we ran a qualitative study with $2$ professional and $1$ amateur game designers. All designer participants went through the following protocol: (1) We first allowed the designers to play the game to get familiar with its mechanics. (2) We then gave them a quick tutorial about the tool and how to use it. (3) We asked designers to use the system to create a playable level. (4) We asked the designers to communicate their level design goal and then use the tool to achieve it. (5) We finally asked the designers to complete a questionnaire related to their experience using the tool.


\subsection{Observing the Designers}
All three designers played the first level shown in the online port of Lode Runner~\footnote{\url{https://loderunnerwebgame.com/game/}} to get familiar with the core game mechanics. After they were comfortable with the game, we run them through a quick tutorial of the tool and then asked them to create their levels (as per our experimental protocol). Throughout this process, we observed them while using the tool and recorded all their interactions with Lode Enhancer. We noticed that all three participants ended up modifying the 16x16 canvas most of the time. 
That was expected as the goal was to end up with a 16x16 playable level.

\begin{figure}[!tb]
    \centering
    \begin{subfigure}[t]{0.32\linewidth}
        \centering
        \includegraphics[width=0.95\textwidth]{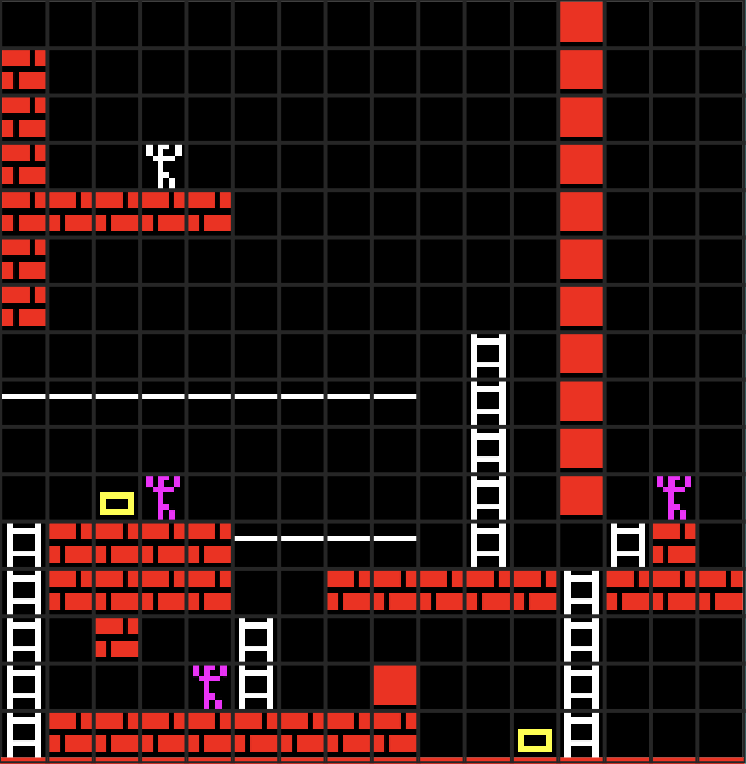}
        \caption{}
        \label{fig:first_1}
    \end{subfigure}
    \begin{subfigure}[t]{0.32\linewidth}
        \centering
        \includegraphics[width=0.95\textwidth]{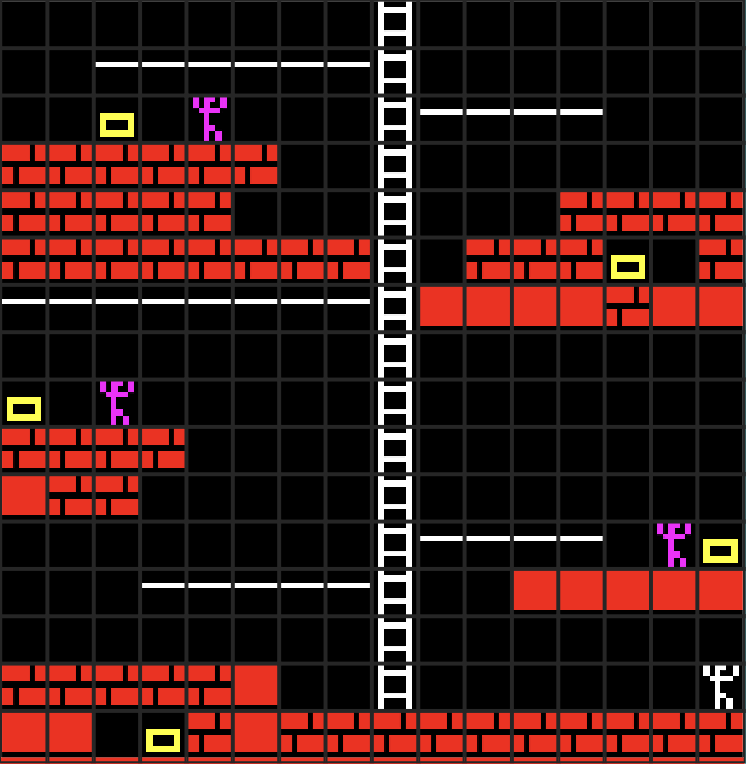}
        \caption{}
        \label{fig:first_2}
    \end{subfigure}
    \begin{subfigure}[t]{0.32\linewidth}
        \centering
        \includegraphics[width=0.95\textwidth]{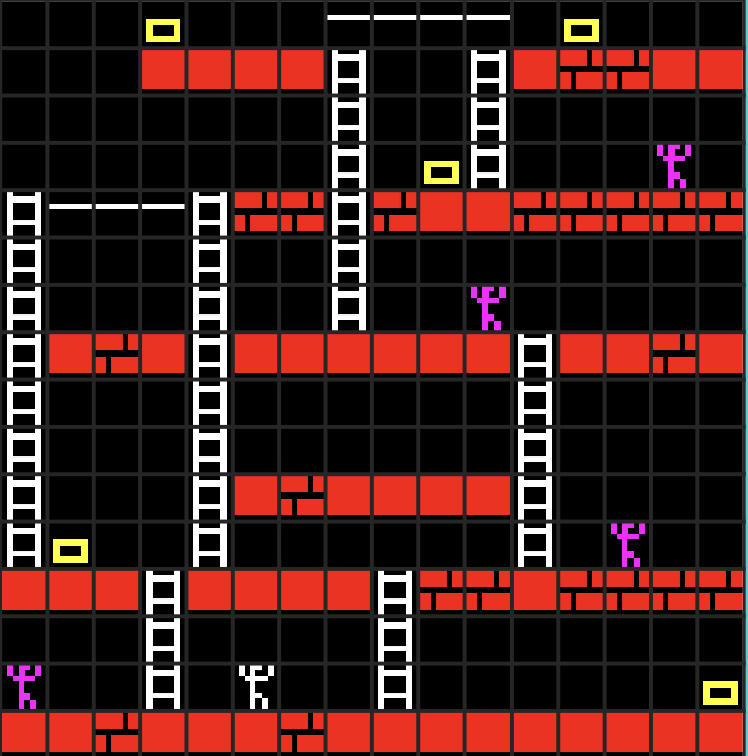}
        \caption{}
        \label{fig:first_3}
    \end{subfigure}
    \caption{Three playable levels designed by 3 different participants when asked to design a playable level.}
    \label{fig:first_levels}
\end{figure}

\begin{figure}[!tb]
    \centering
    \begin{subfigure}[t]{0.32\linewidth}
        \centering
        \captionsetup{width=.9\linewidth}
        \includegraphics[width=0.95\textwidth]{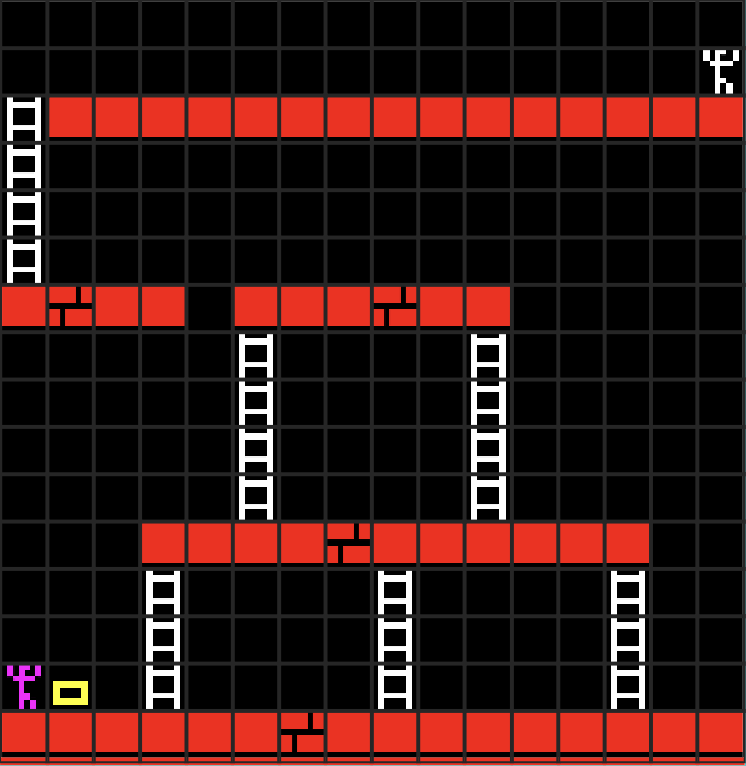}
        \caption{}
        \label{fig:second_1}
    \end{subfigure}
    \begin{subfigure}[t]{0.32\linewidth}
        \centering
        \captionsetup{width=.9\linewidth}
        \includegraphics[width=0.95\textwidth]{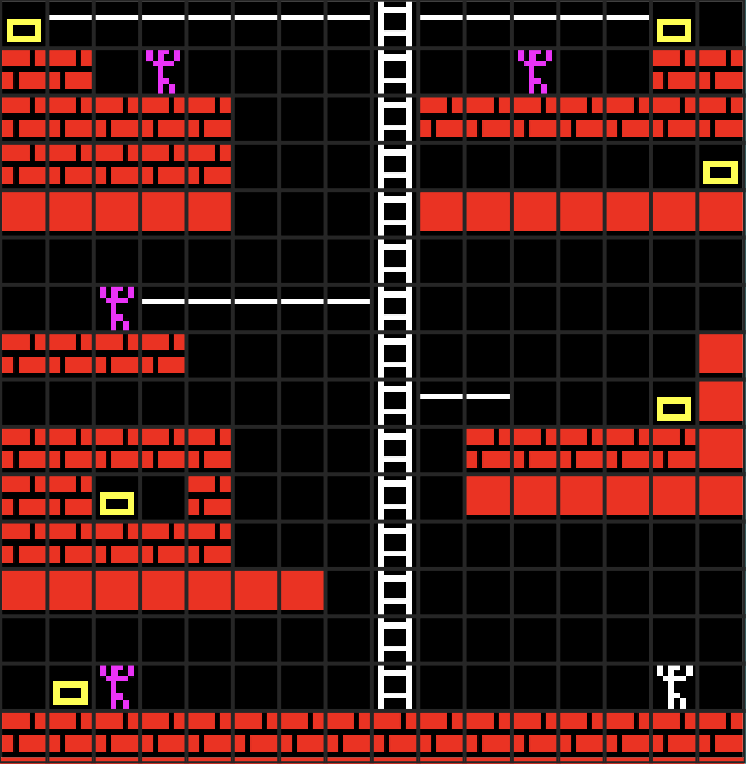}
        \caption{}
        \label{fig:second_2}
    \end{subfigure}
    \begin{subfigure}[t]{0.32\linewidth}
        \centering
        \captionsetup{width=.9\linewidth}
        \includegraphics[width=0.95\textwidth]{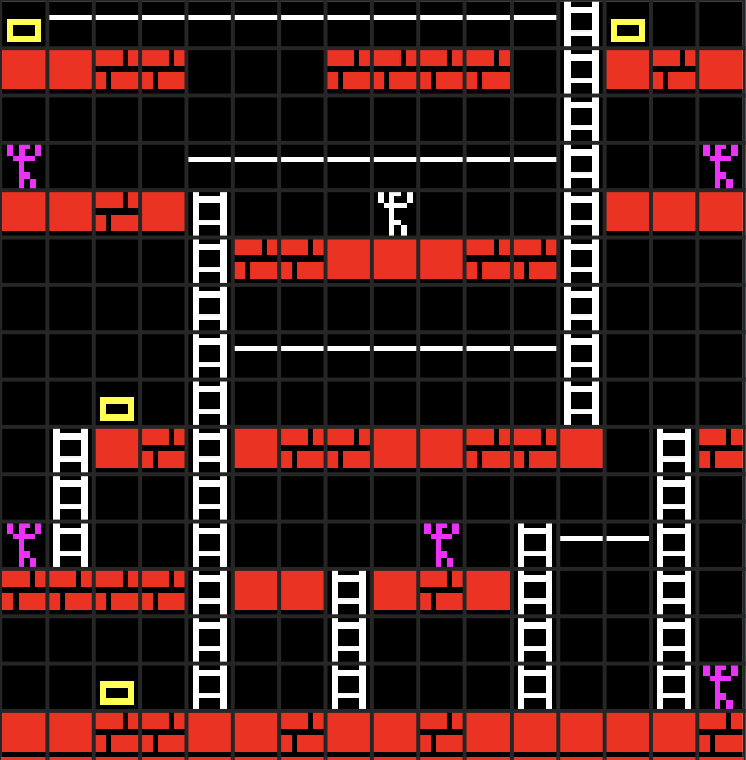}
        \caption{}
        \label{fig:second_3}
    \end{subfigure}
    \caption{Three goal-oriented levels designed by 3 participants. The goals were set by the participants and communicated to us. The goals are (a) create a level with one gold that the enemy will carry and the only way to win is to trap the enemy to drop the gold, (b) create a level with at least 3 enemies and 5 golds where one of the gold is trapped, and (c) create a challenging level with multiple possible paths.}
    \label{fig:second_levels}
\end{figure}

Figure~\ref{fig:first_levels} shows the 3 different playable levels created by the designers without communicating any intent. In this task, the first designer (figure~\ref{fig:first_1}) completely ignored the small and middle canvas and started directly manipulating the 16x16 canvas. Figure~\ref{fig:second_levels} shows the 3 different playable levels that were created after designers communicated a certain goal. In this task, all three designers started working with a small-scale canvas to get an interesting shape in the bigger canvas and then completed their levels by modifying the 16x16 canvas. All participants managed to create a level that they are satisfied with. 

Another observation is that the third designer (figure~\ref{fig:first_3} and~\ref{fig:second_3}) took twice as much time to design their level compared to the others. We assume that the designer was focusing on having a highly detailed level (a lot of solid and brick tile mix) on the canvas, they need to clean some of the up-scaled structures and think about the locations of each brick tile. For all created levels except figure~\ref{fig:second_3}, the designers ignored the 4x4 canvas completely. We think due to the small space of the canvas some designers might have felt restricted. Finally, we noticed that they all ignored the persistence function and did not change the default value for it but since none of them went back and forth between different-size canvases, they never needed its functionality.

\subsection{Questionnaire Results}

After participants completed both levels, we asked them a number of questions and analyzed their responses. 

\subsubsection{What did you like about the system?}

Designers liked how sensible the AI edits are. For example, the first designer tried to design an unreachable space but the upscaler made sure it would be reachable by adding holes and/or ladders. They also liked how it helped them start new ideas and distribute the tiles in different areas pretty fast. For example, when adding a small ladder near some platform, the system was able to connect the ladder to the platform. 

\subsubsection{What did you dislike about the system?}

Designers pointed out that the 4x4 grid was too restrictive for design and that is why they avoided using it. Going back and forth between scales was more troublesome than useful especially later in the design process as the AI was suggesting less polished sections. 
They also pointed out that one of the reasons they avoided the scaler is that they are not familiar with the tool so they did not completely understand the relationship between changing a small canvas and the upscaler output.

\subsubsection{How could enhancing/scaling levels improve level designers' workflow and fulfilling their own design goals?}

For this question, all designers agreed that the system helped them save a little bit of time and created a draft version of the level pretty fast. They also think that it would be more useful if the task was to create a huge level like 128x128 while controlling a small space like 16x16.

\subsubsection{How did you use the persistence function?}

Since all designers never went back and forth between different scales, they completely ignored the persistence function. When we asked them about it, they acknowledged that they either forgot about it or did not need it.

\subsubsection{What other AI/ML functionality would you like to see in a tool like this?}

For future functionality, each designer had an interesting insight into what might be useful. One of them wanted the AI to be able to validate the created space and figure out any problems with their design. Another designer wanted to have more control over the up-scaled canvas so there is more than one option for up-scaling. Another suggestion was to provide the user with the ability to teach the upscaler the meaning behind the small section so it can upscale it in the same meaningful way. This suggestion is mostly about making the AI able to recognize the macro-patterns behind a small design and then be able to produce it on a bigger scale, similar to studies by Baldwin et al.~\cite{baldwin2017mixed}, and Dahlskog and Togelius~\cite{dahlskog2013patterns}.

\section{Discussion}

Any machine-learned design assistance tool will rely on and reproduce patterns in the data it was trained on. For some domains, this might present ethical issues, but in the rather abstract domain of Lode Runner, the main concerns are to which extent it reproduces existing levels and constrains the user's creativity. Informal investigations indicate that the upscaling network does not reproduce any recognizable part of existing levels exactly. The ability to edit freely also means that the editor does not constrain the expression of users, although it can of course bias the user in various directions.

One reason that the designers did not engage much with the persistence functionality is apparently that we did not explain it well enough. This is something that needs to be improved on, given that the main reason why designers did not go back and forth between scales while editing is the worry that edits they had already done would be overwritten by the AI. This is exactly what the persistence functionality is supposed to prevent. Clearly, more work is needed on how to make this smooth and intuitive. 

An important takeaway from the qualitative evaluation is that the labor-saving aspects of upscaling only really come into play for large or high-resolution levels. For 16x16 tiles, the effort saved is rather small. Lode Runner might not be the testbed for this, as most Lode Runner levels are not larger than those we use here. 

For our new proposed architecture, our quantitative analysis shows promise in the ability of the network to mimic local patterns compared to traditional architectures. Having a small canvas and upscaling it multiple times allows the system to repair its own mistakes and make sure the generated levels are in distribution. This allows the system to generate bigger levels, which is the main takeaway of the qualitative evaluation. We also think that having a different output for each layer allowed the network to have different computation power for different tiles with minimum domain knowledge (i.e. the order of the tiles). 
Finally, we think that the skip connection allowed the system to learn a good relationship between user input and target output.

While in this study we investigated upscaling as a design assistance functionality in its own right, it is likely that it will be most efficient and useful as part of a multifaceted AI-powered toolset. One functionality that would very likely work well in tandem with the current upscaling (and downscaling) is a repair agent that makes sure that the generated level is playable. Having a repair function and a larger level canvas will likely make the tool more usable and accessible to designers. We believe that the introduced upscaling method has the potential to be a valuable addition to level editing tools similar to diffusion models for image editing tools~\footnote{\url{https://exchange.adobe.com/apps/cc/114117da/stable-diffusion}}.

\section{Conclusion}

We presented Lode Enhancer, a mixed-initiative level design tool based on the concept of \emph{upscaling} as design assistance. Upscaling uses deep neural networks which are trained on artificially downscaled patches of existing Lode Runner level segments, meaning that they have learned common micro- and macro-patterns of Lode Runner game design. These patterns are then exploited when transforming the missing information of a downscaled level to a more information-rich upscaled level. We also proposed a new architecture we name \emph{layer scaling network} and compared it to traditional architectures. We notice that the new architecture manages to capture the local patterns better than the other networks. We also noticed that the trained networks do not allow a small change in the input to cause severe (undesired or unpredictable) changes in the output level. Finally, we tested our tool through a preliminary qualitative study involving 3 game designers. All designers agreed that the tool is very helpful for creating an initial low-res draft which they can refine at a later stage in a co-creative fashion with the underlying AI. They also agreed that this tool will be even more powerful in complex games or larger-scale maps. 

Lode Enhancer is the first mixed-initiative co-creativity \cite{yannakakis2014mixed} prototype that examined and built on the notion of upscaling. We argue that upscaling is a very important task during level design as it can accelerate the drafting process of large levels. As future research steps, we plan to explore how a tool like Lode Enhancer can operate in more complex games with large levels. Also, we consider adding more control on the upscaler module such that the user can have a choice between different versions of the upscaling. Finally, as the levels created usually require some minor edits from the user to make them playable the next version of Lode Enhancer will be equipped with AI agents that can autonomously repair the upscaled levels and turn them into playable ones. 


\bibliographystyle{ACM-Reference-Format}
\bibliography{sample}

\end{document}